\documentclass[letterpaper]{article} 
\usepackage{aaai25}  
\usepackage{times}  
\usepackage{helvet}  
\usepackage{courier}  
\usepackage[hyphens]{url}  
\usepackage{graphicx} 
\usepackage{natbib}  
\usepackage{caption} 
\usepackage{makecell}
\usepackage{algorithm}
\usepackage{algorithmic}

\usepackage{newfloat}
\usepackage{listings}
\DeclareCaptionStyle{ruled}{labelfont=normalfont,labelsep=colon,strut=off} 
\floatstyle{ruled}
\newfloat{listing}{tb}{lst}{}
\floatname{listing}{Listing}
\title{A Hybrid Theory and Data-driven Approach to Persuasion Detection with \\Large Language Models}
\author{
    Written by AAAI Press Staff\textsuperscript{\rm 1}\thanks{With help from the AAAI Publications Committee.}\\
    AAAI Style Contributions by Pater Patel Schneider,
    Sunil Issar,\\
    J. Scott Penberthy,
    George Ferguson,
    Hans Guesgen,
    Francisco Cruz\equalcontrib,
    Marc Pujol-Gonzalez\equalcontrib
}
\author{
    Gia Bao Hoang\textsuperscript{\rm 1},
    Keith J. Ransom\textsuperscript{\rm 1},
    Rachel G. Stephens\textsuperscript{\rm 1}, \\
    Carolyn Semmler\textsuperscript{\rm 1},
    Nicolas Fay\textsuperscript{\rm 2},
    Lewis Mitchell\textsuperscript{\rm 1}
}
\affiliations{

    \textsuperscript{\rm 1}University of Adelaide\\
    \textsuperscript{\rm 2}University of Western Australia\\
    
    \{giabao.hoang, keith.ransom, rachel.stephens, carolyn.semmler, lewis.mitchell\}@adelaide.edu.au, \\ nicolas.fay@uwa.edu.au
}

\begin{document}

\maketitle

\begin{abstract}
Traditional psychological models of belief revision focus on face-to-face interactions, but with the rise of social media, more effective models are needed to capture belief revision at scale, in this rich text-based online discourse. Here, we use a hybrid approach, utilizing large language models (LLMs) to develop a model that predicts successful persuasion using features derived from psychological experiments. 


Our approach leverages LLM generated ratings of features previously examined in the literature to build a random forest classification model that predicts whether a message will result in belief change. Of the eight features tested, \textit{epistemic emotion} and \textit{willingness to share} were the top-ranking predictors of belief change in the model. Our findings provide insights into the characteristics of persuasive messages and demonstrate how LLMs can enhance models of successful persuasion based on psychological theory. Given these insights, this work has broader applications in fields such as online influence detection and misinformation mitigation, as well as measuring the effectiveness of online narratives.
\end{abstract}

%

\section{Introduction}
Changing someone’s opinion is a key aspect of communication, with broad implications for both individual decisions and societal outcomes. From political campaigns to marketing strategies and daily interactions, understanding how opinions change has been the subject of extensive research, (see e.g., \citealp{cialdini2007influence, dillard2002persuasion, petty2012communication, reardon1991persuasion}). With the rise of online social platforms, interpersonal persuasion can be observed and enacted on a massive scale, not just in direct conversation but also through text-based communication. As \citet{fogg_mass_2008} highlights, debate and argumentation are increasingly shifting to digital spaces. This introduces new complexities in understanding how persuasive messages are created, received, and how they influence the revision of beliefs in online environments.

Persuasion in these environments is particularly important to understand given global concerns about phenomena such as misinformation, polarization, and echo chambers (see e.g., \citealp{arechar_understanding_2023, barbera_social_2020,weber2022promoting}). In addition, influence campaigns are shown to be strategically coordinated efforts designed to shape public opinion using social diffusion processes and media exposure \citep{hwang_social_2012, panagopoulos_campaign_2012}. Therefore, early detection and a deeper understanding of how these processes unfold, along with identifying the characteristics of successful persuasion, are crucial. 

Belief revision is a complex process. Beyond the content of a message, outcomes depend on how individuals engage with information, evaluate its credibility, and respond emotionally. Epistemic emotions (such as curiosity, confusion, and surprise) motivate deeper cognitive engagement \citep{muis_role_2018}, and have been shown to facilitate persuasion when they prompt active message processing \citep{brinol_history_2012}. In parallel, emotional valence influences how information is processed, shaping attention, memory, and judgment \citep{lerner_emotion_2015}. These internal responses often determine whether persuasion occurs, yet they are not always evident from surface-level linguistic features. This complexity underscores the need for models that incorporate psychological insight, not just observable text patterns.



While foundational psychological models like the Elaboration Likelihood Model (ELM) \cite{petty_elaboration_1986} and the Heuristic-Systematic Model (HSM) \cite{chaiken_heuristic_1980} have shaped our understanding of persuasion, they are primarily descriptive and offer limited utility for predicting persuasive success at the level of individual messages. Separately, researchers have also attempted to use machine learning and large language models (LLMs) to analyze persuasion at scale (\citealp{tan_winning_2016, bai_artificial_2023, salvi_conversational_2024}). While promising, these approaches often function as black boxes, offering little insight into how persuasive effects are achieved. As a result, the relationship between linguistic features and psychological theories of persuasion remains difficult to interpret and apply in practice.


In this work, we demonstrate how psychological theories can be combined with LLMs’ statistical representation of language, to enhance existing machine learning architectures to predict persuasion outcomes in text-based discussions. This study utilizes two datasets: the Winning Arguments dataset \citep{tan_winning_2016} for real-world online persuasion analysis, and the Truth Wins dataset \citep{fay_truth_2024}, which provides controlled, human-annotated data from experiments on persuasive and attention-seeking messages. Our focus is on classifying successful versus unsuccessful persuasive attempts in the Winning Arguments dataset, using features derived from the Truth Wins dataset. Our final model leverages a Random Forest architecture trained on LLM-generated persuasion ratings, inspired by psychological experiments. The model predicts whether a message will lead to successful persuasion. Additionally, we examine which features are the most influential predictors within the classification model.

\section{Related Work}



The ELM and the HSM are foundational theories in persuasion research, both proposing dual routes to attitude change, either through deep, content-focused processing or more superficial cues (\citealp{petty_elaboration_1986, chaiken_heuristic_1980}). While influential, these models are largely descriptive, and offer little practical guidance for predicting persuasive success at the level of individual messages. Their application often requires adaptation to specific contexts, topics, and communication media, which limits their scalability in computational settings.

In response, researchers have explored ways to make persuasion more measurable. \citet{zhao_measure_2011} proposed the Perceived Argument Strength (PAS) scale as a subjective measure of persuasive quality, but it relies on self-reports and lacks scalability. \citet{youk_measures_2024} developed the Measures of Argument Strength (MAS), identifying linguistic features -- such as citations, abstractness, and moral language -- that are associated with higher perceived persuasiveness in large-scale online debates. \citet{lukin_argument_2017} further emphasized how personality traits shape responses to emotional versus factual arguments, highlighting the importance of audience adaptation. While these studies move toward operationalizing persuasion, they remain constrained by subjective judgement, context sensitivity, handcrafted features, and limited predictive power -- motivating the further shift toward data-driven approaches, discussed next.

Online platforms like Change My View (CMV) subreddit have become an important place to study persuasion at scale. In CMV, users influence opinions via textual exchanges. Two important features of CMV that are valuable for computational work are the {\em delta symbol}, awarded by a user to the reply that changed their view, and the {\em karma score}, which reflects community endorsement through up-votes and down-votes. While deltas serve as explicit markers of persuasive success, karma is often used as a proxy for argument quality. In addition, as the forum is highly moderated, comments are usually argumentative, containing reasoning rather than \textit{ad hominem} attacks or other form of online bullying. 

\citet{tan_winning_2016} conducted a large-scale analysis of CMV (the Winning Arguments dataset), focusing on delta award predictions, based only on delta given by the original poster (OP). They examined the interaction dynamic associated with successful attempts at persuasion, and built a logistic regression model based on language factors and linguistic style. The model was able to predict delta changes effectively, but struggled with reliably distinguishing stance malleability (whether an OP will change their position), indicating difficulty in reliably distinguishing stance shifts and highlighting the complexity of persuasion in online discourse. Subsequent studies have built on this dataset using more advanced techniques. \citet{dutta_changing_2020} introduced an attention-based hierarchical Long Short-Term Memory model to classify persuasive conversations on CMV by analyzing argument-specific components, such as claims and premises. Similarly, \citet{wei_is_2016} used karma scores as a proxy for success, finding that argumentative features outperformed surface-level cues like length and punctuation in predicting persuasiveness. These works demonstrate the promise of data-driven persuasion modeling, but also expose its dependence on platform-specific feedback, limiting broader applicability.

Recent work has shifted from predicting persuasive outcomes to detecting persuasive strategies, often using deep learning and transformer-based NLP models. \citet{karki_using_2022} classified Cialdini’s six persuasion principles (Reciprocation, Consistency, Social Proof, Likeability, Authority, and Scarcity) in phishing emails using machine-learning transformers based on BERT. In a related direction, \citet{wang_persuasion_2020} developed a personalized persuasive dialogue system by integrating sentence-level and contextual features, including turn position, sentiment, and character embeddings, enabling the classification of persuasion strategies in conversations. While these models perform well, they remain largely opaque and task-specific, offering limited insight into underlying persuasive mechanisms and weak connections to psychological theory, underscoring the need for models that are both interpretable and cognitively grounded.


\begin{figure*}[t]
    \centering
    \includegraphics[width=0.9\linewidth]{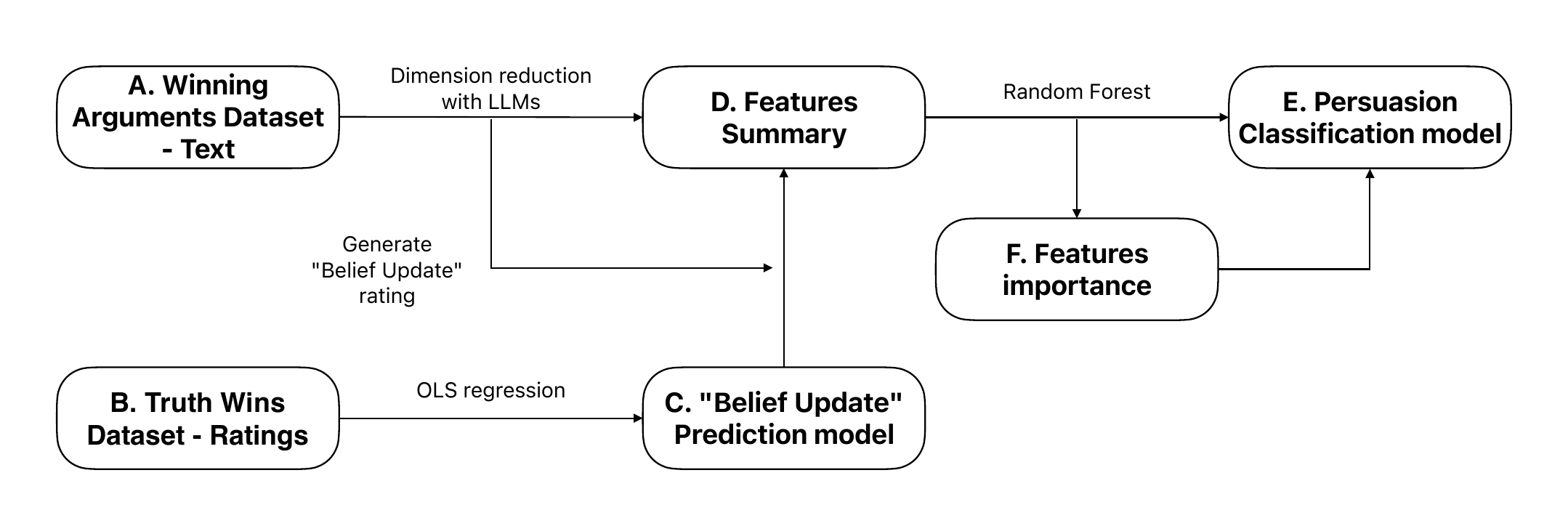}
    \caption{Flowchart showing the main steps of our hybrid model approach.}
    \label{fig:persuasion-model}
\end{figure*}


Despite recent advances, traditional methods for persuasion detection continue to face challenges related to scalability, generalization, and personalization, often relying on large annotated datasets and lengthy text inputs. Large language models (LLMs) offer a promising alternative, leveraging their statistically grounded representations of human language to overcome these limitations. However, LLMs introduce new challenges -- most notably, a lack of transparency. Operating as black-box systems, they are often difficult to interpret, which limits their usefulness in contexts requiring explainability. The current research project addresses these gaps by building on and extending three key areas of prior work: (1) models for predicting persuasive success, (2) studies identifying persuasive strategies and message characteristics, and (3) applications of LLMs in persuasion contexts. This integrative approach supports real-time applications such as influence campaign detection and misinformation mitigation, while also enabling greater flexibility in identifying and analyzing the persuasive elements within language.

\section{Experimental Setup}
Figure \ref{fig:persuasion-model} outlines the workflow of our hybrid model for persuasion classification in target data such as the Winning Arguments dataset. The goal is to predict whether a response successfully changes the original poster’s view by leveraging interpretable, theory-driven features in combination with scalable language model outputs. We begin by identifying a set of psychologically grounded features from the Truth Wins dataset (detailed below), including Attention, Interesting-If-True, Positive Emotion, Negative Emotion, Influential, Shareable, and Truthfulness (Figure~\ref{fig:persuasion-model}.B). These features form the basis for a structured representation of persuasive content. 

Next, comments from the Winning Arguments dataset (Figure~\ref{fig:persuasion-model}.A) are passed to LLMs, which extract feature ratings based on the features derived from the Truth Wins dataset (Figure~\ref{fig:persuasion-model}.D). To estimate the likelihood of belief change, we train a separate belief update regression model on annotated belief shift data from Truth Wins (Figure~\ref{fig:persuasion-model}.C). 

Together, the feature ratings generated by the LLMs and the belief update scores predicted by the regression model were input into a Random Forest classifier for binary classification (Figure \ref{fig:persuasion-model}.E). Performance of the Random Forest classifier is then measured using the test set of Winning Arguments. Using the Random Forest model’s mechanism for evaluating feature importance enables the removal of noise or negatively contributing features, further refining the final model (Figure \ref{fig:persuasion-model}.F). However, while this refinement did little to improve accuracy, it did allow us to simplify the model and clarify the ranking of features.

\subsection{Datasets}
\subsubsection{Winning Arguments Dataset (Figure \ref{fig:persuasion-model}.A)}
This is the primary dataset for this study, sourced from the Change My View subreddit where users invite others to challenge their opinions by presenting counterarguments \citep{tan_winning_2016}. Persuasion is explicitly marked by the OP awarding a delta symbol ($\Delta$) to comments that successfully change their view, thus distinguishing between successful persuasion and unsuccessful comments.

We focus on root replies, the first-level responses from challengers to the OP. For each post, we extract one positive (successful) and one negative (unsuccessful) comment. Since posts often receive multiple challenges, we control for content similarity by selecting comment pairs with the highest Jaccard similarity scores, in an effort to ensure that comparisons are based more on persuasiveness rather than content differences. The dataset consists of 3,456 posts for training and 807 for testing, each including the OP’s text, a positive reply, and a negative reply. This offers a robust and ecologically valid foundation for studying persuasive dynamics in online discourse.

\subsubsection{Truth Wins Dataset (Figure \ref{fig:persuasion-model}.B)}
The Truth Wins dataset was collected from a series of human-subject experiments designed to study persuasion and attention-seeking in written messages across diverse topics \citep{fay_truth_2024}. Participants were asked to generate messages either to persuade or to garner online attention. A second between-subjects manipulation varied whether participants were told to base their messages on true information, false information, or on any combination they wished. 

Each message was rated by ten independent human annotators along multiple dimensions including: belief update (ranging from -100 to 100, reflecting the change between prior versus post belief score), attention, interesting, interesting-if-true, positive emotion, negative emotion, perceived effectiveness of influence, shareability, and truthfulness. Except for belief update, all ratings were recorded on a 5-point Likert scale: "not at all", "slightly", "somewhat", "very much" and "extremely". Following \citet{fay_truth_2024}, we assigned each item an integer value from 1 ("not at all") to 5 ("extremely"), and aggregated the ratings by averaging across raters to reduce individual variance while improving signal stability and generalizability. This aggregated structure aligns with our focus on identifying consistent predictors in successful persuasion.

These features are grounded in prior research. Epistemic emotions (interesting, interesting-if-true) promote deeper processing and belief change \citep{muis_role_2018, brinol_history_2012}. Interestingness boosts engagement, though its effect on persuasion is context-dependent \citep{alwitt_effects_2000, murphy_examining_2005}. Shareability captures perceived social attention, which supports deeper analysis and persuasive impact \citep{shteynberg_broadcast_2016, shteynberg_collective_2018}. Emotional valence shapes attention and persuasion; positive emotions enhance message acceptance, while relevant negative emotions can also increase persuasion \citep{lerner_emotion_2015, petty_emotion_2015, huntsinger_flexible_2016}. Truthfulness reflects perceived accuracy, which makes messages more persuasive and less likely to be dismissed. \citep{chaiken_heuristic_1980, eagly_psychology_1993, kunda_case_1990, petty_elaboration_1986, tormala_role_2016}.

While the original study explored the influence of truthfulness on persuasion and attention-seeking, our research examines broader factors affecting persuasiveness, acknowledging that attention-seeking messages can also carry persuasive elements. The controlled, human-labeled data from the Truth Wins can be used to help classify the more ecologically valid discourse found in the CMV-based Winning Arguments dataset, offering a comprehensive view of persuasive mechanisms across different contexts.

\subsection{Baseline Models}
To evaluate our proposed hybrid model, we compare it to three representative baselines: theory-driven inference, transparent surface-level text modeling, and black-box reasoning via large language models (LLMs). Together, these baselines form a spectrum of interpretability, linguistic abstraction, and data reliance.

\subsubsection{Theory-Driven Model -- Belief Update Regression (Figure \ref{fig:persuasion-model}.C)}
As a baseline, we modeled persuasion as belief change. Using the Truth Wins dataset, which includes human-annotated belief update ratings, we trained an ordinal least square (OLS) regression model to identify significant features predictive of belief update, using the other eight key features studied by \citet{fay_truth_2024}. Following a stepwise procedure, variables were added or removed based on p-value thresholds (0.05), with problematic features excluded to ensure model stability.

The resulting model was then applied to the Winning Arguments dataset to generate predicted belief update scores for each comment. These scores were then thresholded -- tuned for accuracy on the training set -- to assign binary persuasion labels. Though not optimized for classifying persuasive success, it offers a theory-informed perspective based on belief update scores. In our hybrid model, these belief update scores are also used as a feature, contributing to a psychologically grounded signal that complements language-based predictions. Note that belief update and successful persuasion are distinct constructs: not all persuasive responses result in explicit belief change, and not all belief changes are externally acknowledged.

\subsubsection{Transparent Language Model — Logistic Regression on Term Frequencies}
This baseline uses a shallow, interpretable language model to classify persuasion based solely on surface-level lexical features. Each comment is tokenized and vectorized using term frequency (TF), omitting inverse document frequency (IDF) to better accommodate short-form text, where rare words may not carry more informational weight. We train a logistic regression with 10-fold cross-validation, tuned on regularization strength, and evaluated performance using ROC AUC. This model provides a transparent lexical benchmark, revealing how much signal can be captured from word-level frequency patterns alone.

\subsubsection{Black-Box Language Model — Zero-Shot LLM Classification}
To assess the potential of large language models (LLMs), we evaluated three models: LLaMA3-70B (LLaMA3), Gemma2-9B (Gemma2), and Mixtral-8x7B (Mixtral) in a zero-shot persuasion task. Each model receives the OP's post and two responses: one that successfully persuaded the OP and one that did not. The models are then prompted to select the response they judge to be more persuasive based on the input. The full prompt is provided in Listing~\ref{lst:llm_zeroshot} in Appendix~A. All models are run with temperature set to 0 and a fixed seed (42) to ensure high probability responses and reproducibility. 

This approach demonstrates the scalability and adaptability of LLMs, which can be applied without fine-tuning or additional supervision. Although tools like Gemma Scope provide interpretability scaffolding, these models still operate with a high degree of opacity, and their predictions can be difficult to interpret or validate -- especially in high-stakes contexts such as misinformation detection, where transparency and accountability are critical. To mitigate these limitations, we subsequently use the same LLMs to generate structured feature ratings. These outputs are integrated into a hybrid model to support more interpretable and feature-aware classification of persuasive language. 

\subsection{Hybrid Model}
\begin{figure*}[t]
    \centering
    \includegraphics[width=\linewidth]{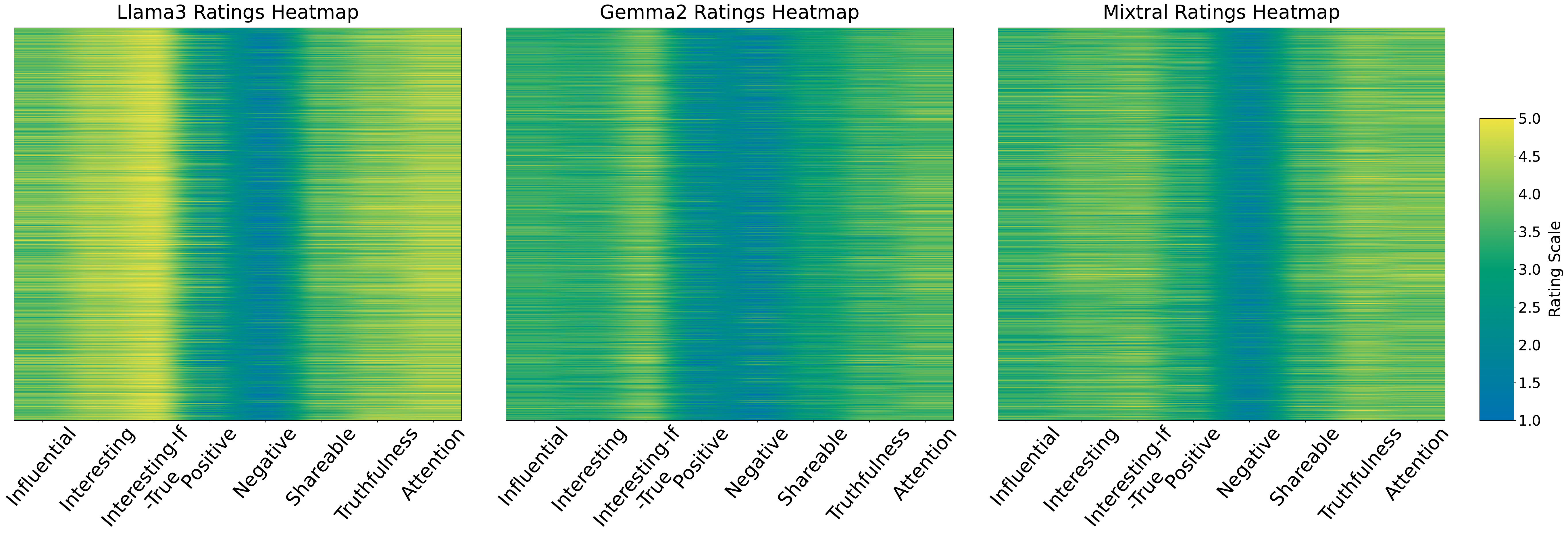}
    \caption{LLMs Generated Ratings Heatmap - This visualizes the ratings assigned by each LLM for each analysed feature (x-axis) for each argument in the Winning Arguments dataset (y-axis).} 
    \label{fig:heatmaps}
\end{figure*}

\begin{table*}[t]
    \centering
    \begin{tabular}{lccccc}
        \hline
        \textbf{Model} & \textbf{Data-Driven} & \textbf{Theory-Driven} & \textbf{Hybrid Independent} & \textbf{Hybrid Interaction} \\
        \hline
        Transparent (Logistic regression) & 56.50\% & - & - & - \\
        Black-Box (LLaMA3) & 64.93\% & 54.30\% & 73.17\% & 73.11\% \\
        Black-Box (Gemma2) & 56.90\% & 63.80\% & 82.32\% & 82.69\% \\
        Black-Box (Mixtral) & 61.71\% & 58.20\% & 72.55\% & 73.17\% \\
        \hline
    \end{tabular}
    \caption{Accuracy comparison of models used for classifying successful persuasion (positive) vs. unsuccessful (negative) comments in the Winning Arguments dataset. \textbf{Transparent Data-Driven} refers to a logistic regression on term frequencies. \textbf{Black-Box Data-Driven} represents zero-shot classification from black-box LLMs. \textbf{Theory-Driven} uses thresholded belief update scores derived from regression models. \textbf{Hybrid Independent} and \textbf{Hybrid Interaction} refer to Random Forest models that combine features with belief update scores, using either independent terms only or including pairwise feature interactions, respectively.} 
    \label{tab:models_accuracy}
\end{table*}

\begin{figure*}[t]
    \centering
    \includegraphics[width=0.68\linewidth]{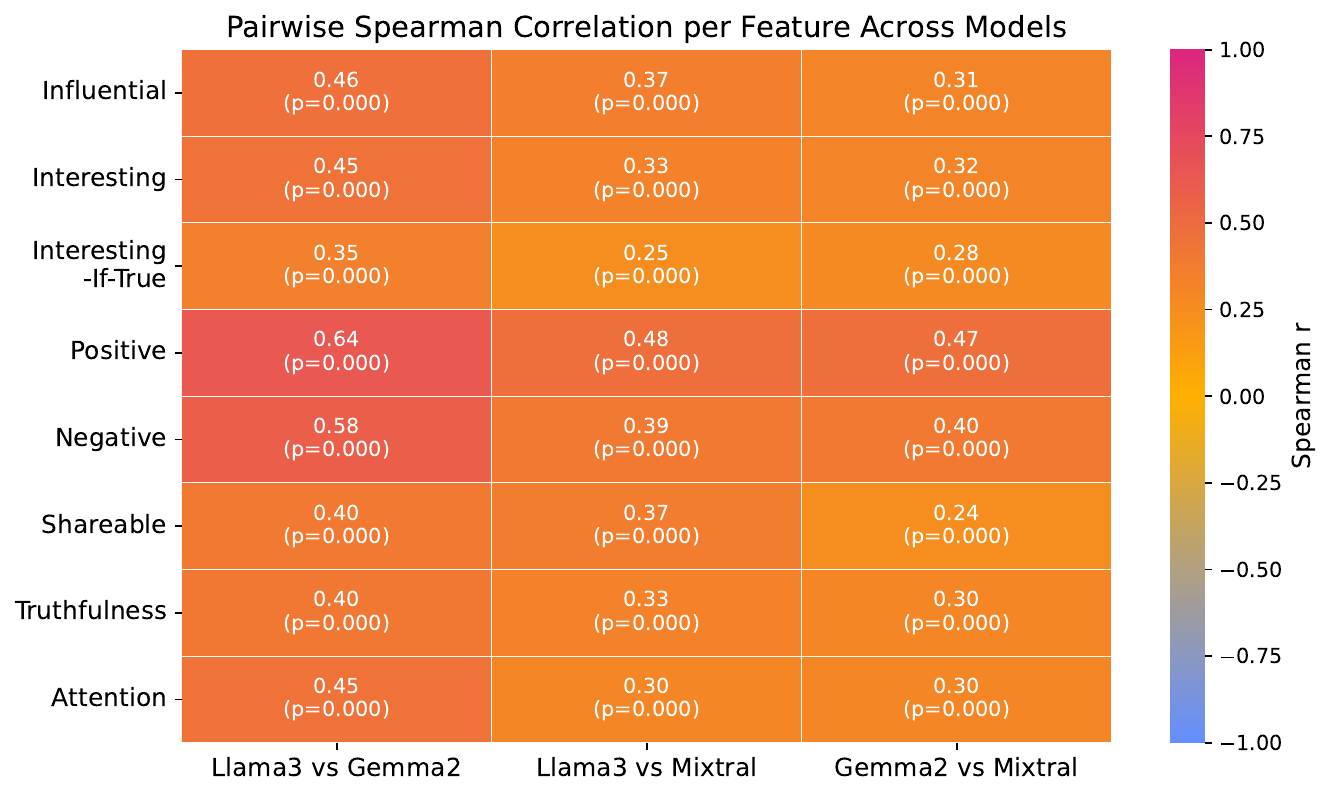}
    \caption{Spearman Rank Correlation between LLMs' generated ratings. Each cell represents the correlation between the ratings generated by the given pair of LLMs (x-axis) for a given feature (y-axis).}
    \label{fig:pairwise_corr}
\end{figure*}

\subsubsection{Feature Extraction Using LLMs (Figure \ref{fig:persuasion-model}.D)}  
Using the same three LLMs used for the Black-Box model  (LLaMA3, Gemma2, and Mixtral), we generated feature-by-feature ratings for comments in the Winning Arguments dataset. Given the original post along with a pair of comments (one labeled positive, one negative), each model was prompted to generate ratings for both comments in relation to the original post. The full prompt is provided in Listing~\ref{lst:llm_extraction} in Appendix~A. The selected features rated by the LLMs were adapted from the Truth Wins dataset and are defined as follows:

\begin{enumerate}
    \item \textit{Influential}: How effectively the comment influences the reader’s perspective.
    \item \textit{Interesting}: How engaging and attention-grabbing the comment is.
    \item \textit{Interesting-If-True}: Potential interest level if the comment’s claims were true.
    \item \textit{Positive}: Overall positivity, focusing on an uplifting or encouraging tone.
    \item \textit{Negative}: Overall negativity, considering critical or discouraging aspects.
    \item \textit{Shareable}: Likelihood of the comment being shared.
    \item \textit{Truthfulness}: Apparent accuracy and credibility of the comment.
    \item \textit{Attention}: Ability to capture and maintain reader focus.
\end{enumerate}

To avoid ambiguity, we replaced the original \citet{fay_truth_2024} feature label \textit{Persuasive} with \textit{Influential}. This \textit{Influential} feature captures the perceived effectiveness of a comment in influencing a reader’s perspective, rather than directly measuring actual belief change. This distinction prevents confusion between the general notion of perceived “persuasiveness” as a feature and the task of predicting successful persuasion itself. Using the feature schema derived from the Truth Wins dataset, we obtained LLM-generated ratings for the same features in the Winning Arguments dataset, providing a theory-driven input layer for persuasion classification. This conceptualization aligns closely with the probabilistic framework described by \citet{wyer_jr_probabilistic_1970}, which distinguishes between a message’s influence on a premise and the subsequent belief update. This approach enabled a structured analysis of persuasive strategies in real-world discourse, grounded in a controlled, theory-informed framework.

\subsubsection{Independent vs. Interaction Terms}  
To assess the role of feature interactions in persuasion prediction, we developed two hybrid model variants: one using only independent features, and another incorporating feature interactions. The independent-term model evaluates each feature's contribution individually. In contrast, the interaction-term model includes pairwise combinations of features alongside their independent counterparts. For simplicity, interactions are limited to two-feature pairs. This approach enables us to determine whether interactions between features significantly enhance predictive performance or whether individual features alone are sufficient for effective persuasion classification.

\subsubsection{Random Forest Classification (Figure \ref{fig:persuasion-model}.E)}
We use a Random Forest (RF) classifier \citep{breiman_random_2001} for its strong predictive performance combined with the interpretability of decision trees. Our hybrid RF models are trained on LLM-generated features and belief update scores derived from regression models. The hybrid models are then evaluated on the test set. At this stage, the OP's post content is not taken into consideration. The RF-independent model combines LLM-generated features with belief update scores derived from the independent-term regression, while the RF-interaction model extends this by incorporating pairwise interaction features and scores from the interaction-term regression. Using RF’s built-in permutation-based Variable Importance (VI), we assess which characteristics extracted by LLMs most meaningfully predict persuasive success, driving model performance.

VI assesses each feature by measuring decrease in classification accuracy when its values are randomly permuted, directly quantifies each feature’s predictive contribution. Throughout this study, unless otherwise specified, “feature importance” specifically refers to VI. Using VI, we can remove noisy or redundant features, streamline the model, and mitigate potential overfitting without sacrificing accuracy. Although VI can be biased toward high-cardinality features \citep{strobl_bias_2007}, our dataset consists exclusively of ordinal categorical features (rated 1 to 5) and continuous belief update scores, mitigating this bias. Our random forest classifier uses 300 trees, balancing complexity and generalizability. A fixed random seed (42) ensures reproducibility.


\section{Results}
\subsection{Cross-Model Agreement in Generated Feature Ratings}


To examine how different language models interpret identical instructions for feature ratings, we compared ratings generated by LLaMA3, Gemma2, and Mixtral on the full Winning Arguments dataset. Each model independently rated the following features: \textit{Influential}, \textit{Interesting}, \textit{Interesting-If-True}, \textit{Positive}, \textit{Negative}, \textit{Truthfulness}, \textit{Attention}, and \textit{Shareable}.

Heatmaps (Figure~\ref{fig:heatmaps}) reveal systematic rating differences across models. LLaMA3 consistently produced higher ratings for features such as \textit{Influential}, \textit{Interesting}, \textit{Interesting-If-True}, and \textit{Attention}, suggesting a relatively generous interpretation of these criteria. In contrast, Gemma2 and Mixtral tended toward lower ratings overall, with Gemma2 notably rating \textit{Positive} lower and giving higher ratings for \textit{Interesting-If-True} and \textit{Attention}. Mixtral’s ratings typically fell between those of LLaMA3 and Gemma2.

A pairwise correlation analysis (Figure~\ref{fig:pairwise_corr}) demonstrated consistent positive and statistically significant (p-value $<$ 0.001) monotonic correlations between models, despite differences in absolute ratings. LLaMA3 and Gemma2 consistently showed the strongest correlations across features, with values reaching 0.64 for \textit{Positive} and 0.58 for \textit{Negative}, highlighting close agreement in valence judgments. In contrast, strong predictive features like \textit{Interesting-If-True} showed weaker correlations across all pairs, pointing to greater ambiguity or model-specific interpretation. 

These differences in both absolute scores and correlation structure help explain variation in downstream performance across models. At the same time, the consistent alignment in rating trends between LLMs, as shown in Figure~\ref{fig:pairwise_corr} through statistically significant positive monotonic correlations, indicates a degree of consistency across models. This observed alignment suggests that the LLM-generated features may be suitable for use in downstream modeling, despite variation in individual rating scales.


\subsection{Model Accuracy Performance}
We evaluated model performance for classifying successful persuasion (positive) vs. unsuccessful (negative) comments in the Winning Arguments dataset, summarized in Table \ref{tab:models_accuracy}. 

Hybrid classifiers integrating LLM-generated features outperformed both the theory-driven model, and transparent and black-box language model baselines. The Gemma2-based hybrid model achieved the highest accuracy, roughly 82\% for both independent-term and interaction-term variants, approximately a 25\% improvement over its black-box baseline. Hybrid models built on LLaMA3 and Mixtral followed closely, with accuracies around 73\% and similarly small differences between the two feature configurations.

Interestingly, including feature interactions and removing negatively contributing features had little impact on predictive accuracy across all hybrid models. This indicates that independent features, as generated by LLMs, already sufficiently capture the predictive signals necessary for effective persuasion classification.

\subsection{Feature Importance (Figure \ref{fig:persuasion-model}.F)}
\begin{figure*}[t]
    \centering
    \begin{minipage}{0.32\linewidth}
        \centering
        \includegraphics[width=\linewidth]{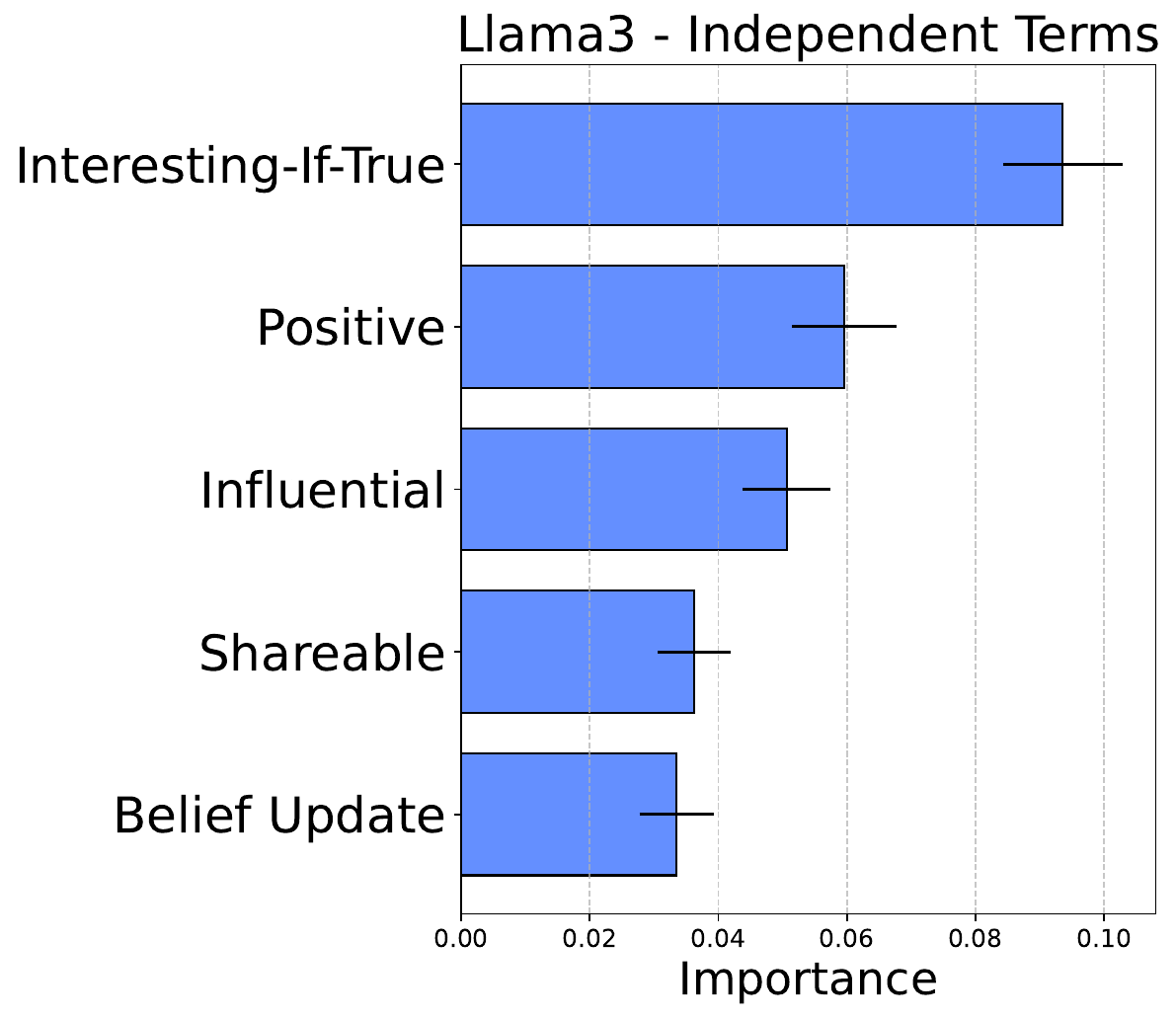}
    \end{minipage} \hspace{0.01\linewidth}
    \begin{minipage}{0.32\linewidth}
        \centering
        \includegraphics[width=\linewidth]{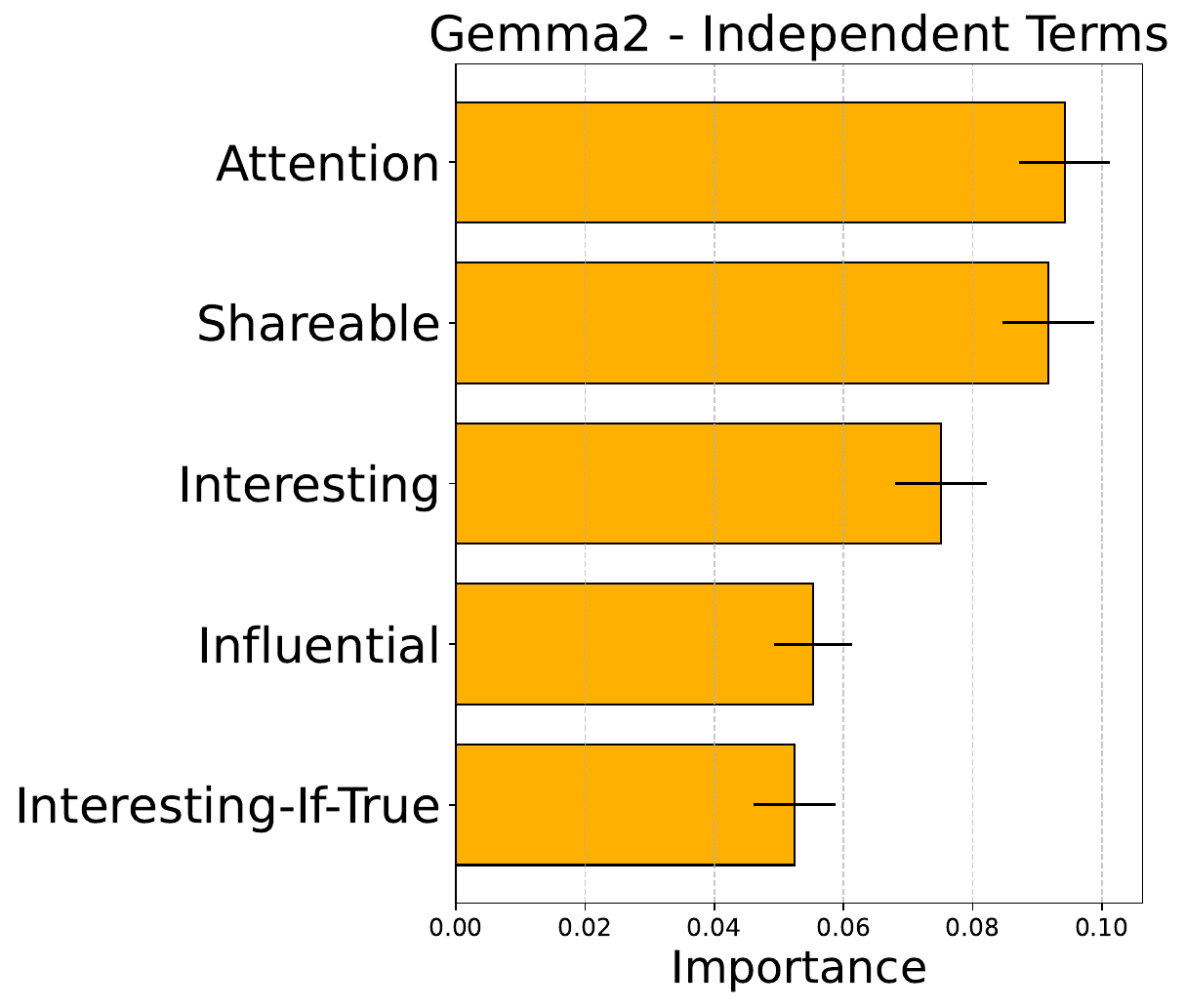}
    \end{minipage} \hspace{0.01\linewidth}
    \begin{minipage}{0.32\linewidth}
        \centering
        \includegraphics[width=\linewidth]{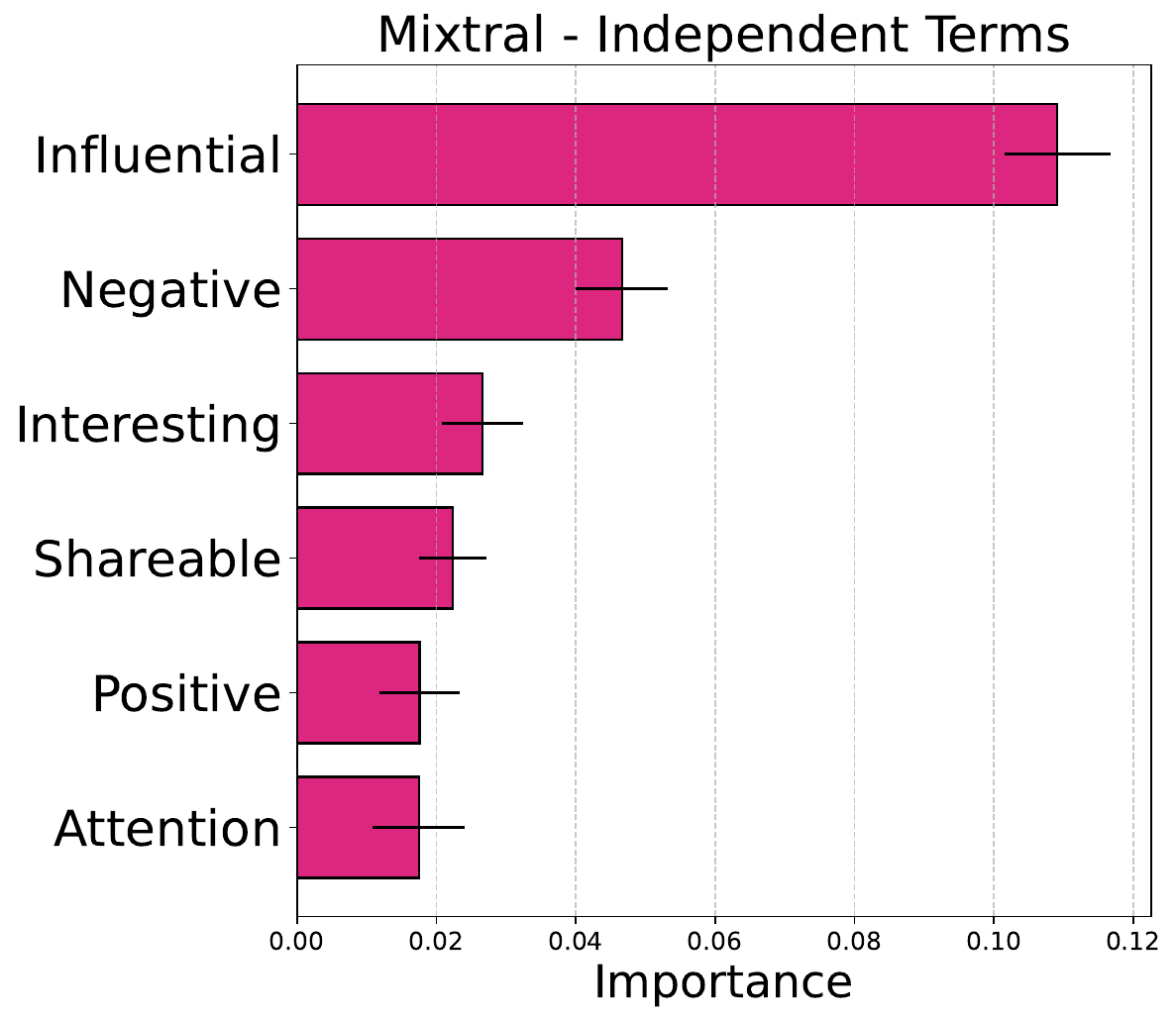}
    \end{minipage}

    \vspace{0.0em}  

    \begin{minipage}{0.32\linewidth}
        \centering
        \includegraphics[width=\linewidth]{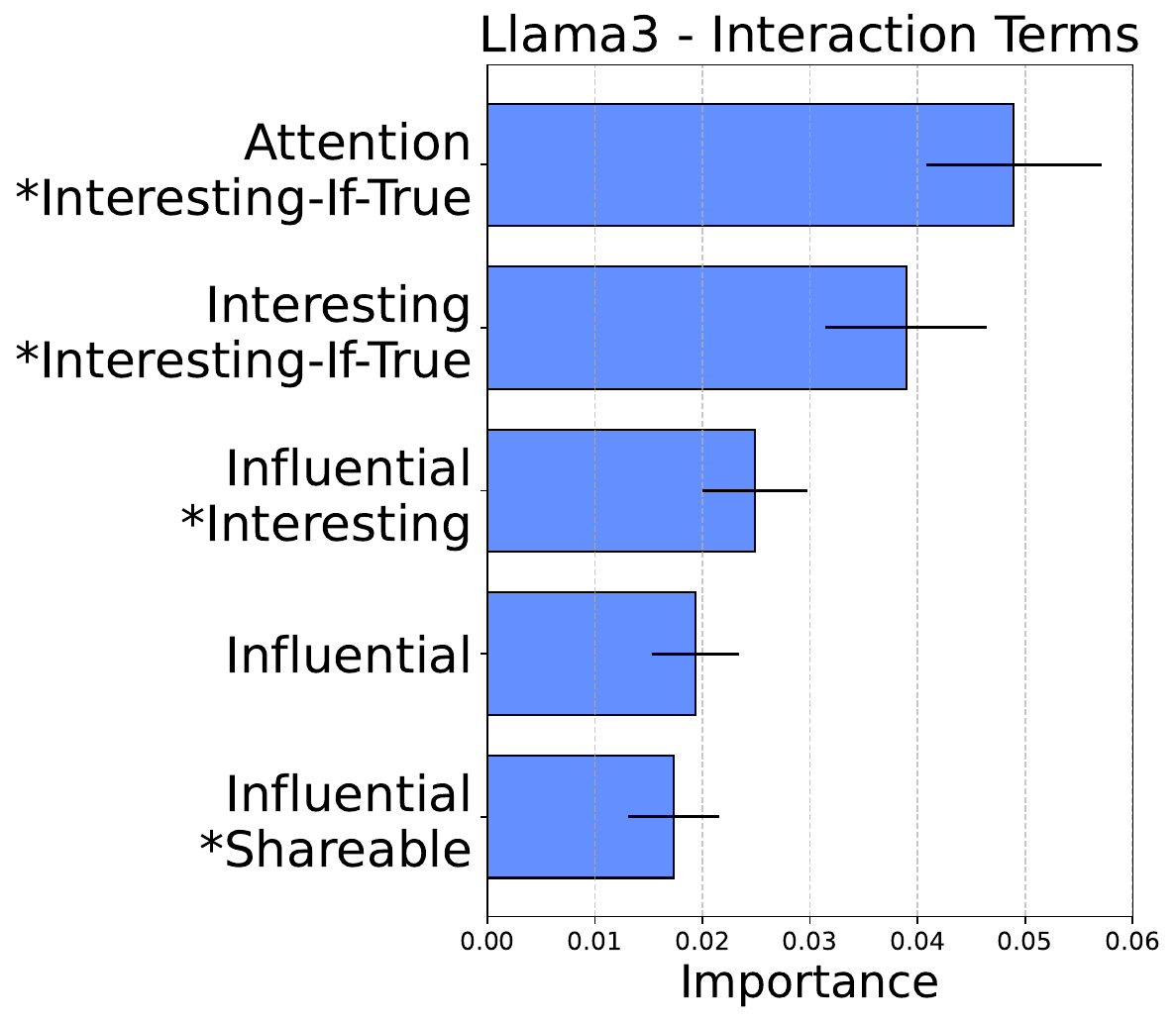}
    \end{minipage} \hspace{0.01\linewidth}
    \begin{minipage}{0.32\linewidth}
        \centering
        \includegraphics[width=\linewidth]{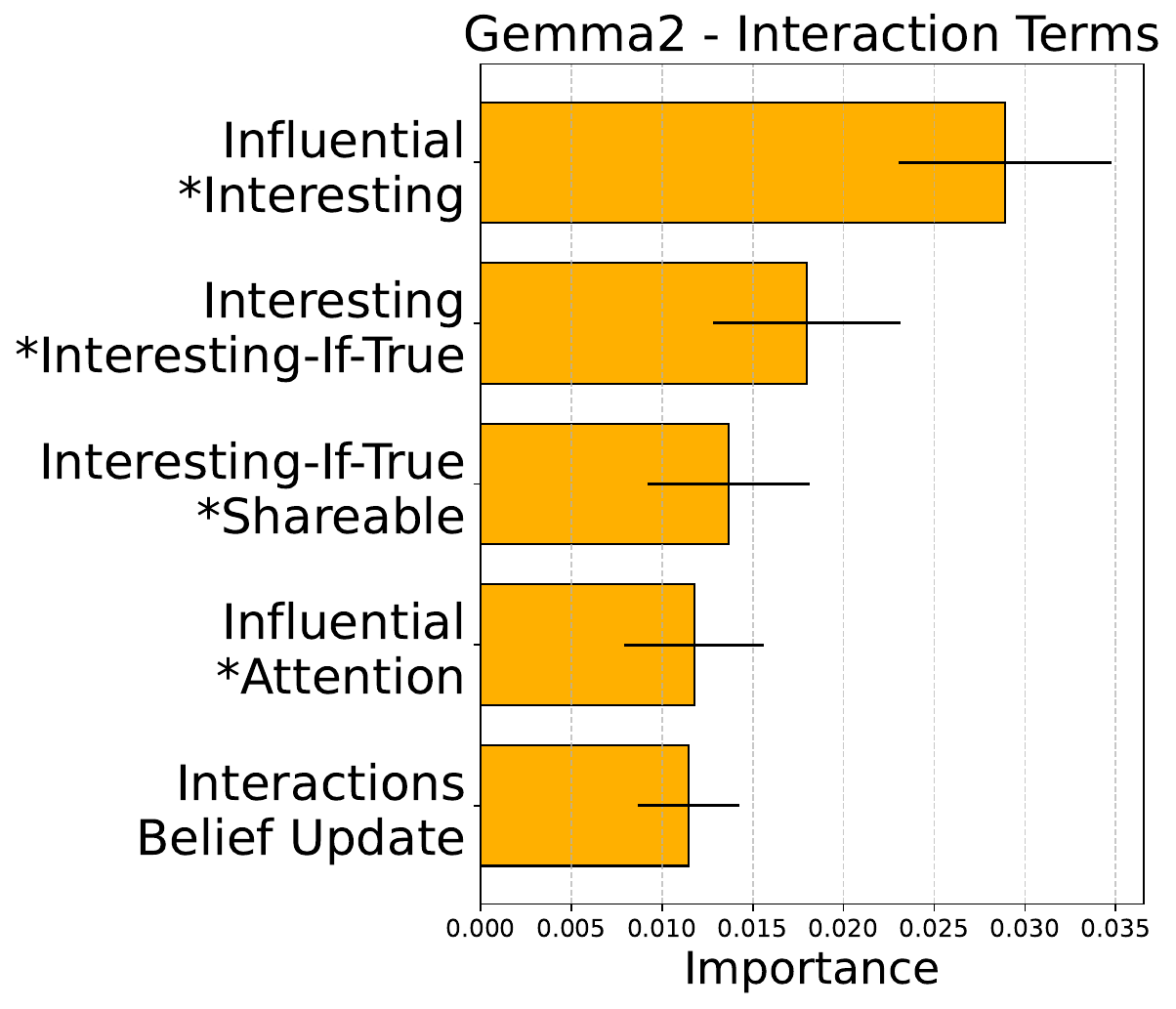}
    \end{minipage} \hspace{0.01\linewidth}
    \begin{minipage}{0.32\linewidth}
        \centering
        \includegraphics[width=\linewidth]{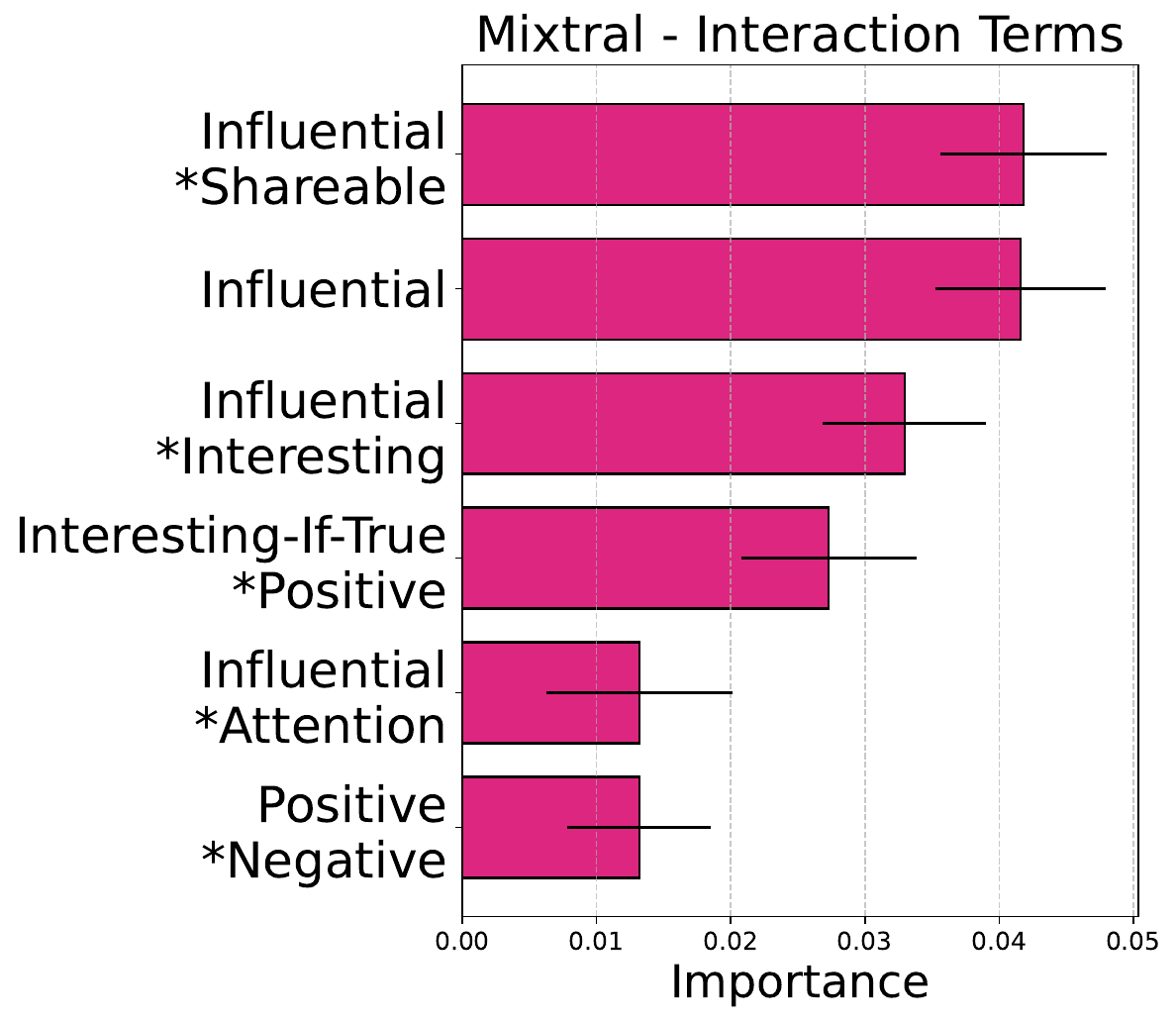}
    \end{minipage}

    \caption{Top 5 Most Important Features in the Hybrid Model (Random Forest), with Independent Terms shown on the top row and Interaction Terms on the bottom row. Each bar represents the mean feature importance across 100 permutations, and the error line indicates the standard deviation. For the Mixtral-based models, six features are shown due to a tie in importance scores for the fifth position.}
    \label{fig:permutation}
\end{figure*}

Figure~\ref{fig:permutation} summarizes the most important model features. Despite differences amongst LLMs, all consistently identify a core set of important features: \textit{Influential}, \textit{Shareable}, \textit{Interesting/Interesting-If-True}, and \textit{Attention}. Specifically, in the independent-term models, LLaMA3 emphasized \textit{Interesting-If-True}, Gemma2 prioritized \textit{Attention}, and Mixtral strongly highlighted \textit{Influential}. Interaction-term models revealed that combinations involving \textit{Interesting-If-True}, \textit{Interesting}, and \textit{Influential} substantially enhanced predictive performance, indicating potentially informative interactions among these features. Additionally, \textit{Belief Update} was influential in Gemma2, suggesting its relevance in predicting persuasive success. Emotional features (\textit{Positive} and \textit{Negative}) showed inconsistent predictive power, warranting further investigation. Notably, \textit{Truthfulness} consistently ranked among the least important features, which may reflect limited variation in the accuracy of information within CMV posts rather than a lack of relevance to persuasion.

\section{Discussion}



Overall, our findings show that hybrid models outperform both theory-driven and data-driven baselines, suggesting that combining LLMs with structured, theory-informed features improves the prediction of persuasive success. Notably, LLMs perform best when used to generate interpretable ratings for individual features, highlighting the value of integrating statistical language models with domain knowledge to better model persuasive success.

\subsection{Theory-Driven Model}
The belief update prediction model, trained on the Truth Wins dataset and applied to LLM-generated feature ratings from the Winning Arguments dataset, reveals critical insights into cross-domain generalizability. While the model offers a broad indication of belief change, its simplicity -- using ordinal linear regression -- fails to capture the complexities of successful persuasion. Incorporating the belief update score into the final model adds some interpretability but does not significantly improve performance. This suggests that belief update scores, while useful, are not sufficient to fully capture the dynamics of persuasive impact.

What we learn here is that, although theory-driven models can provide valuable insights, they struggle to translate across domains and fail to account for the nuanced, dynamic nature of persuasion. This underscores the need for more sophisticated, hybrid models that combine theoretical insights with the strengths of machine learning to better capture the complexities of persuasion and improve generalizability across diverse data sources.

\subsection{Transparent Data-Driven Model}
The logistic regression model based on term frequency (TF) highlights the limitations of using word frequency alone for persuasion detection. TF models, particularly for short-form text like CMV comments, struggle to capture linguistic and contextual elements such as sentiment and interaction dynamics. This emphasizes the need for more theory-informed features and predictors that connect specific linguistic cues to successful persuasion -- an area where the hybrid model, integrating both theory-driven insights and data-driven techniques, can offer a more comprehensive solution.

\subsection{Black-Box Data-Driven Model - Leveraging LLMs}
Zero-shot classifiers, based on LLMs, demonstrated an improvement over logistic regression, though directly prompting LLMs to select the more persuasive comment proved less effective. LLMs excel at converting linguistic features into ratings that capture statistical regularities in human communication. This ability to represent language numerically enhances the performance of downstream models, particularly when structured data is needed. Although some models performed better than others in zero-shot classification, the real value of LLMs lies in their capacity to encode linguistic features into structured data. This underscores the potential of LLMs to aid in tasks requiring large-scale, feature-rich representations.

Research highlights the potential of LLMs as a powerful tool for persuasion analysis and belief influence. Studies have shown that LLMs, such as GPT-3.5, align closely with human judgments on persuasiveness, with high correlation between human and LLM-generated ratings \citep{fay_truth_2024}. In fact, LLMs have been found to outperform humans in assessing persuasiveness across various topics and demographics \citep{salvi_conversational_2024} and can effectively influence opinions on both polarized and less polarized issues \citep{bai_artificial_2023}. Additionally, \citet{costello_durably_2024} demonstrated that conversations with GPT-4 Turbo could reduce belief in conspiracy theories by roughly 20\%, with effects lasting up to two months, showcasing LLMs' potential in belief change.

Given these capabilities, future research should explore comparisons between LLM-generated and human ratings for accuracy and consistency. Hybrid approaches could further enhance performance in complex tasks like influence campaign detection and misinformation analysis, where a deeper understanding of persuasive content and context is crucial.

\subsection{Cognitive Features and Biases in Persuasion}
The top predictive features identified across all three models were \textit{Interesting-If-True}, \textit{Influential}, and \textit{Shareable}. In the context of the CMV forum, where the primary goal is to change the OP's viewpoint, these features become particularly relevant. Unlike everyday conversations, which may not always prioritize persuasion, the CMV forum encourages high-quality, persuasive arguments. As a result, less obvious features like curiosity (\textit{Interesting}-If-True) and social sharing (\textit{Shareable}) become crucial factors in shifting opinions.

In the Hybrid Independent models, \textit{Interesting-If-True} ranks highest for LLaMA3, while \textit{Attention} holds the top position for Gemma2. Both features also rank among the top five for Mixtral. These features are closely tied to epistemic emotions, which stimulate cognitive engagement \cite{muis_role_2018}. Curiosity fosters exploration and information seeking, encouraging individuals to gather evidence and critically evaluate messages to resolve their curiosity \citep{vogl_surprise_2019}. In the context of persuasion, a message that sparks curiosity -- such as “That’s interesting, is it true?” -- may prompt the audience to scrutinize it more rigorously, potentially verifying its credibility before accepting it.

Similarly, interest sustains attention and systematic processing. As part of the broader category of epistemic emotions, interest -- like curiosity and surprise -- has been shown to enhance message processing and facilitate persuasion by prompting deeper cognitive engagement \citep{muis_role_2018, brinol_history_2012}. These emotions are also associated with increased message sharing and heightened perceptions of credibility, even when the underlying content lacks truthfulness \citep{rijo_thats_2023}. This may explain why \textit{Shareable} is influential across all three models, alongside \textit{Interesting-If-True}. Both enhance emotional engagement, mediating how information is perceived, evaluated, and disseminated online.

\textit{Truthfulness} was a weaker predictor of persuasive success in CMV. Perceived truthfulness might be more important in environments with a wider mix of objectively true and false information. Research by \citet{rijo_thats_2023} suggests that fake news often elicits stronger epistemic emotional responses than true news, making it appear more credible and leading to wider dissemination. These findings highlight the dominance of perceived truth over objective truth in shaping belief formation and information spread. Moreover, regarding misinformation, \citet{pennycook_nudging_2022} highlighted that while people can typically distinguish between true and false information when asked, this analytical distinction fades in informal contexts like social media. In these environments, social and affective factors -- rather than truth-based analysis -- primarily drive sharing behavior.

Given this focus on persuasion rather than truthfulness, future research could further explore the role and drivers of perceived truth quality in the Winning Arguments dataset to assess whether it plays a significant role in persuasion. For example, \citet{fazio_knowledge_2015} demonstrated the illusory truth effect in news classification: repetition of a claim (even if false) can increase its perceived truth and likelihood of being shared. Investigating whether similar effects apply in the CMV context could offer further insights into how persuasion works, regardless of objective truth.

Emotion was also a significant feature, with \textit{Positive} emotions ranking highly for LLaMA3 and \textit{Negative} emotions for Mixtral. This aligns with research on how emotions influence persuasion. According to \citet{naranowicz_mood_2022}, emotional state affects critical engagement: negative moods promote scrutiny of argument quality, while positive moods encourage reliance on credibility cues and general trust. Furthermore, \citet{paletz_emotional_2023} and \citet{fay_truth_2024} found that positive emotions drive content sharing, reinforcing the role of emotional engagement in persuasion. These findings highlight the importance of emotional features in our model.

While our analysis captures emotional valence, it overlooks arousal, which plays a crucial role in persuasion. High-arousal emotions -- whether positive or negative -- amplify persuasive impact by encouraging fast, intuitive processing \citep{luhring_emotions_2024}, leading to snap judgments and reduced source verification. Since arousal is absent from our feature set, future research should examine its role in persuasion. Incorporating arousal-related features could provide deeper insights into decision-making, credibility assessments, and content sharing, offering a more comprehensive understanding of emotional engagement in persuasion.


\section{Conclusion}
This study demonstrates the potential of large language models (LLMs) as effective tools for predicting successful persuasion in online discourse. By generating feature ratings on dimensions such as influence effectiveness, interest, and shareability, LLMs provide a scalable method for analyzing persuasive language, bridging the gap between manual annotation and automated analysis.

Beyond their practical applications, the findings support theoretical perspectives on persuasion. Feature importance analysis in the Random Forest model revealed significant patterns aligning with existing literature, reinforcing the role of linguistic and cognitive factors in persuasive content. This supports the validity of the feature set and highlights its relevance in computational persuasion research.

Trained on high-quality arguments from Change My View subreddit, the model represents a step forward in both theoretical and applied research. Automating the rating process for belief prediction offers valuable insights into opinion shifts in digital spaces. By integrating theory-driven insights with data-driven modeling, this work advances both our understanding of persuasive mechanisms and real-world applications, including influence campaign detection, misinformation mitigation, and content moderation.

\section*{Limitations}
The Winning Arguments dataset, sourced from a forum focused on persuasion, may limit the generalizability of our findings. Since users on CMV are actively trying to change others’ opinions, the dynamics may differ from more organic or varied settings. Future research should apply the model to other datasets where persuasive intent is less explicit, such as general social media platforms, news comments, or product reviews, to evaluate whether the model performs well across different online environments.



Another key limitation of this study is the potential bias introduced by relying on LLM-generated ratings. While LLMs offer a scalable way to assess textual dimensions, their consistency and accuracy relative to human annotations remain uncertain. Prior work has shown that LLMs can exhibit preference for specific token patterns \citep{zheng_large_2024}, produce less consistent ratings than human annotators \citep{stureborg_large_2024}, and reflect cognitive biases learned from training data \citep{dillion_can_2023}. These issues can affect tasks that require nuanced or objective judgment. Although LLMs may align with human judgments in broad patterns, their outputs can still be biased or unstable, potentially impacting downstream modeling performance.

A related concern is the lack of validation against human-sourced ratings. The hybrid models in this study rely primarily on LLM-generated ratings for the classification task, but the absence of human annotations limits our ability to assess the reliability of those inputs. Future work could compare LLM-generated ratings with human-annotated counterparts for the same features. This would provide a form of sanity check, establish a ground truth for model evaluation, and help identify systematic rating discrepancies across models.


Relying on LLM-generated feature ratings may overlook important linguistic nuances such as grammar, sentence structure, and subtle rhetorical or emotional cues, which are critical for understanding persuasive success. Future work could incorporate more advanced natural language processing techniques, integrating syntactic and semantic analysis alongside LLM-generated features to capture a deeper, richer understanding of persuasion.

LLMs often function as black boxes, making their internal decision-making opaque. This creates challenges in understanding how they determine which features of an argument are persuasive. Future research should explore interpretability techniques to better understand the patterns and factors that drive the model's outputs. These tools can provide insights into the patterns LLMs use to generate persuasive judgments, enhancing model transparency and usability.

\section*{Ethical Consideration}
This study draws upon two previous datasets: the Winning Arguments dataset and the Truth Wins dataset. Both datasets were collected in prior research and are used here under the assumption that the original studies adhered to appropriate ethical standards for data collection. Readers are encouraged to refer to the respective publications for further details on data sourcing, consent, and anonymization procedures.

A portion of the data in this study is generated using Large Language Models (LLMs). While these tools offer efficiency and scale, they also raise several ethical concerns. Most notably, LLMs function as black boxes. Users have limited insight into the data used during training, which may contain biases, or misinformation. Furthermore, the proprietary nature of the LLMs employed (accessed via the Groq Cloud API) means they are not open-source, and their behavior or availability can change without notice. These models are costly to retrain and may become outdated over time. Additional risks include the potential for output degradation, loss of coherence, and generation of repetitive or harmful content. These limitations must be taken into account when interpreting results involving LLM-generated data.

The development of a persuasion prediction model presents a dual-use dilemma. On one hand, this research aims to enhance our understanding of persuasive communication, with the broader goal of fostering safer and more constructive online environments. By identifying linguistic and structural features associated with successful persuasion, we hope to contribute to the creation of more responsible and effective messaging strategies. On the other hand, we recognize that these insights could be misused—to manipulate public opinion, spread misinformation, or exploit vulnerable audiences. As such, we stress that the findings of this study must be applied with caution and responsibility. Ethical safeguards, transparency in deployment, and continuous monitoring are essential to mitigate potential misuse.

\section*{Acknowledgements}
Lewis Mitchell acknowledges support from the Australian Government through the Australian Research Council’s Discovery Projects funding scheme (project DP210103700). Nicolas Fay acknowledges support from the Office of National Intelligence and Australian Research Council (project NI210100224).

\bibliography{final}

\appendix
\section{LLMs Prompts}
\begin{lstlisting}[caption={Prompt used for zero-shot classification with LLMs.}, label={lst:llm_zeroshot}]
<|begin_of_text|><|start_header_id|>system<|end_header_id|>
You are a classifier evaluating the persuasiveness of two comments on an original post. 
Identify which comment is more persuasive. 
Label the more persuasive comment as "positive" (1) and the less persuasive as "negative" (0).

Input includes:
- "op_text": original post text
- "comment1": first comment
- "comment2": second comment

Provide the result as a JSON object for each comment. No explanation is required.
<|eot_id|><|start_header_id|>user<|end_header_id|>
Here is the context and comments: {context}
<|eot_id|><|start_header_id|>assistant<|end_header_id|>
\end{lstlisting}

\begin{lstlisting}[caption={Prompt used for LLM-based feature extraction. In the original prompt, the label \textit{Persuasive} was used; however, we refer to it as \textit{Influential} throughout the paper to avoid ambiguity, as discussed in Section 3.3}, label={lst:llm_extraction}]
begin_of_text|><|start_header_id|>system<|end_header_id|>
You are an expert evaluator tasked with rating two comments based on their context within an original post. 
Each feature is rated from "one" (lowest) to "five" (highest).

Input includes: "op_text" (original post), "comment1" (first comment in response to op_text), and "comment2" (second comment in response to op_text).

Features to evaluate:
- Persuasive: How effectively the comment influences the reader's perspective.
- Interesting: How engaging and attention-grabbing the comment is.
- Interesting if True: Potential interest level if the comment's claims were true.
- Positive: Overall positivity, focusing on an uplifting or encouraging tone.
- Negative: Overall negativity, considering critical or discouraging aspects.
- Shareable: Likelihood of the comment being shared.
- Truthfulness: Apparent accuracy and credibility of the comment.
- Attention: Ability to capture and maintain reader focus.

Return the ratings for each feature for both comments as a JSON object with keys for each feature under "comment1" and "comment2" based on their relation to "op_text," using words ("one" to "five") for ratings. 
Provide only the JSON object with no explanation.

<|eot_id|><|start_header_id|>user<|end_header_id>
Here is the context and comments: {context}
<|eot_id|><|start_header_id|>assistant<|end_header_id>
\end{lstlisting}

\end{document}